\documentclass{article}

\usepackage{microtype}
\usepackage{graphicx}
\usepackage{subfigure}
\usepackage{booktabs} 
\usepackage{enumitem} 
\newcommand{\R}{\mathbb{R}}
\usepackage{hyperref}
\usepackage{arydshln}
\usepackage{enumitem}
\usepackage[ruled,vlined]{algorithm2e}

\usepackage[accepted]{icml2025}

\usepackage{amsmath}
\usepackage{amssymb}
\usepackage{mathtools}
\usepackage{amsthm}
\usepackage[capitalize,noabbrev]{cleveref}

\theoremstyle{plain}

\theoremstyle{definition}

\theoremstyle{remark}

\usepackage[textsize=tiny]{todonotes}

\icmltitlerunning{Locality-aware Surrogates for Gradient-based Black-box Optimization}

\begin{document}

\allowdisplaybreaks  
\setlength{\abovedisplayskip}{3pt} 
\setlength{\belowdisplayskip}{3pt} 
\setlength{\abovedisplayshortskip}{3pt}
\setlength{\belowdisplayshortskip}{3pt} 

\twocolumn[
\icmltitle{Locality-aware Surrogates for Gradient-based Black-box Optimization}

\icmlsetsymbol{equal}{*}

\begin{icmlauthorlist}
\icmlauthor{Ali Momeni$^*$}{sai,epfl}
\icmlauthor{Stefan Uhlich}{ssseu}
\icmlauthor{Arun Venkitaraman}{sai,sssj}
\icmlauthor{Chia-Yu Hsieh}{sai,sssj}
\icmlauthor{Andrea Bonetti}{sai,sssj}
\icmlauthor{Ryoga Matsuo}{ssseu,epfl}
\icmlauthor{Eisaku Ohbuchi}{sssj}
\icmlauthor{Lorenzo Servadei}{sai,sssj}
\end{icmlauthorlist}

\icmlaffiliation{epfl}{EPFL, Switzerland}
\icmlaffiliation{sai}{SonyAI, Switzerland}
\icmlaffiliation{ssseu}{Sony Semiconductor Solutions Europe, Germany}
\icmlaffiliation{sssj}{Sony Semiconductor Solutions, Japan}

\icmlcorrespondingauthor{Ali Momeni}{ali.momeni@epfl.ch}

\icmlkeywords{Surrogate models, Gradient estimation, Physical systems}

\vskip 0.3in
]



\printAffiliationsAndNotice{Work done during an internship at SonyAI.}  


\begin{abstract}
In physics and engineering, many processes are modeled using non-differentiable black-box simulators, making the optimization of such functions particularly challenging.  To address such cases, inspired by the Gradient Theorem, we propose locality-aware surrogate models for active model-based black-box optimization. We first establish a theoretical connection between gradient alignment and the minimization of a Gradient Path Integral Equation (\textit{GradPIE}) loss, which enforces consistency of the surrogate’s gradients in local regions of the design space. Leveraging this theoretical insight, we develop a scalable training algorithm that minimizes the \textit{GradPIE} loss, enabling both offline and online learning while maintaining computational efficiency. We evaluate our approach on three real-world tasks -- spanning automated \textit{in silico} experiments such as coupled nonlinear oscillators, analog circuits, and optical systems -- and demonstrate consistent improvements in optimization efficiency under limited query budgets. Our results offer dependable solutions for both offline and online optimization tasks where reliable gradient estimation is needed.
\end{abstract}

\section{Introduction}
Optimizing black-box objective functions over large design spaces is a major challenge in many scientific and engineering fields. Examples include designing molecules, proteins, drugs, biological sequences, and materials \cite{sarkisyan2016local,nguyen2005evolutionary, si2016high,ashby2000multi}. 
Although several approaches, such as genetic algorithms \cite{banzhaf1998genetic} and Bayesian optimization (BayOpt) \cite{shahriari2015taking}, are commonly used for non-differentiable black-box optimization, we focus on gradient-based optimization due to its scalability and efficiency.
One common approach is to train a differentiable surrogate model on given data to approximate the objective function value (or its inverse) for unknown inputs. Once trained, we can perform gradient ascent on the input space to find the best input points \cite{kumar2020model, brookes2019conditioning, hutter2011sequential}. This \textit{in silico} method, known as offline model-based black-box optimization, simplifies the optimal design problem into a straightforward application of supervised learning and gradient ascent, without actively querying the black-box function during optimization. A key assumption of this approach is that an "accurate" surrogate model can be learned across the entire input space, which is often not possible due to the limited offline training data. The forward model (here a surrogate model) may incorrectly assign high scores to points outside the training data range \cite{krishnamoorthy2022generative}. These inaccurate predictions can mislead the optimization process toward sub-optimal candidates. Several solutions have been proposed, mainly focusing on adding conservative preferences to the search or surrogate training process \cite{trabucco2021conservative,kumar2020model,fannjiang2020autofocused,chemingui2024offline,dao2024boosting}. For example, Conservative Objective Models (COMs) \cite{trabucco2021conservative} address this issue by penalizing high scores for out-of-dataset points, but this can also prevent exploration of high-quality points far from the training data.

\begin{figure*}[t!]
\includegraphics[width=\linewidth,trim=0 35 0 10]{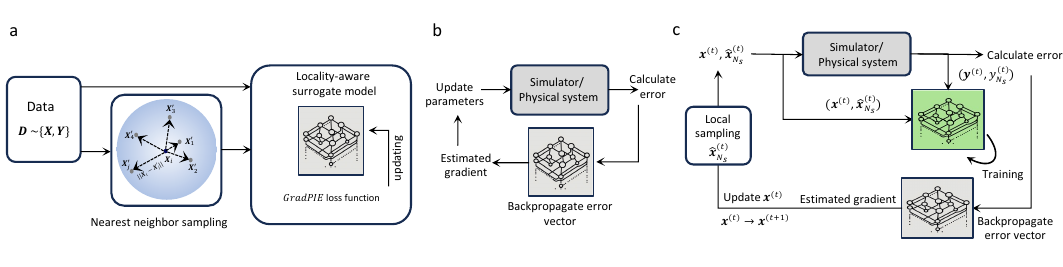}
\caption{\textbf{a} Our method calculates the $k$-nearest neighbors for each sample in the dataset. These neighbors are used to train the locality-aware surrogate model with the \textit{GradPIE} loss function. \textbf{b} Active black-box optimization (ABBO) using the surrogate model trained offline. \textbf{c} ABBO with online training of the surrogate model.}
\label{fig:locality-aware}
\end{figure*}

On the other hand, \textit{active black-box optimization} (ABBO) addresses the issue of unreliable forward model predictions for out-of-distribution offline data by allowing the querying of the black-box function during the forward pass. Recently, stochastic gradient estimators \cite{mohamed2020monte}, such as the REINFORCE estimator \cite{williams1992simple}, have been employed to estimate gradients of non-differentiable functions, enabling gradient-based optimization. To use the strengths of gradient-based optimization and mitigating the high variance often associated with score function gradient estimators, a recent approach involves training a surrogate model as before but using it exclusively in the backward pass to estimate the gradient of the objective function with respect to the inputs \cite{shirobokov2020black,grathwohl2017backpropagation}. This approach is suitable for various black-box optimization problems where automated querying of the black-box function is possible, typically in automated \textit{in silico} or \textit{in situ} experimental settings, such as simulators \cite{shirobokov2020black}, physical systems \cite{wright2022deep}, robotics \cite{de2018end,degrave2019differentiable}, and smart sensors \cite{zhou2020near,mennel2020ultrafast}. Additionally, this technique has recently been experimentally applied to the training of physical neural networks (analog neural networks), as shown in \cite{wright2022deep,zheng2023dual,spall2022hybrid,oguz2024optical,momeni2023backpropagation}. In this case, the overall effectiveness of the optimization loop depends on how accurately the surrogate model can estimate the gradient under limited query budgets. This leads to an important question: how can we train such surrogate models to achieve more accurate gradient estimation within the same query budgets?

In this paper, inspired by the gradient theorem~\cite{WilliamsonTrotter2004}, we propose \emph{locality-aware} surrogate models for ABBO. The main contributions of this work are:

$1-$ To address the question mentioned earlier, we first theoretically demonstrate that minimizing the  \textit{GradPIE} loss —derived from the Gradient Theorem— ensures alignment between the gradients of the surrogate model and the black-box function.  This theoretical result is validated experimentally on a Coupled Nonlinear Oscillator Network, used as a toy example.

$2-$ We propose a scalable training algorithm for locality-aware surrogate models to minimize the \textit{GradPIE} loss. This algorithm supports both, offline training (leveraging pre-collected datasets) and online adaptation (refining gradients iteratively during optimization), ensuring computational efficiency even in high-dimensional design spaces.

$3-$ We evaluate our framework on three real-world tasks —spanning automated \textit{in silico} experiments (e.g., simulators) and physical systems— where automatic querying of black-box functions is feasible. Our results demonstrate consistent improvements in optimization performance under limited query budgets compared to traditional methods.

\section{Background and Problem Setup}
\textbf{Black-box Optimization.} Suppose \( \mathfrak{X} \in \mathbb{R}^{D_i} \) is an input space, and each \( \mathbf{x} \in \mathfrak{X} \) is a candidate input. Let \( \mathbf{F}: \mathfrak{X} \rightarrow \mathbb{R}^{D_o} \) represent the black-box system (simulator or physical system) that maps any input \( \mathbf{x} \in \mathfrak{X} \) to an output \( \mathbf{y} = \mathbf{F}(\mathbf{x}) \). Also, let \( \psi : \mathbb{R}^{D_o} \to \mathbb{R} \) be an objective function that gives a real value. In general, the goal is to find an optimal input \( \mathbf{x}^* \in \mathfrak{X} \) that maximizes the objective function associated with the output of the black-box system \( \mathbf{F}(\mathbf{x}) \), i.e.,
\begin{align}
\mathbf{x}^* \triangleq \arg \max_{\mathbf{x} \in \mathfrak{X}} \psi \big(\mathbf{F}(\mathbf{x}) \big).
\label{eq1}
\end{align}
We are given a dataset consisting of \(N\) input-output pairs $\mathcal{D} = \{(\mathbf{x}_1, \mathbf{y}_1), (\mathbf{x}_2, \mathbf{y}_2), \dots, (\mathbf{x}_N, \mathbf{y}_N)\}$, where each \( \mathbf{y}_i = \mathbf{F}(\mathbf{x}_i) \) is the output of the black-box system \(\mathbf{F}\).

\textbf{Surrogate model (Base-model).} 
We focus on an optimization scheme that uses the black-box system (simulator or physical system) in the forward pass of the optimization loop and a pre-trained surrogate model in the backward pass (see Fig.~\ref{fig:locality-aware}b). The role of the pre-trained surrogate model is to estimate the gradient of the black-box system’s output with respect to the input. 

We can train a surrogate model $\mathbf{\hat F}(\mathbf{x}; \boldsymbol{\theta})$ using the dataset \( \mathcal{D} \) to approximate \( \mathbf{F}(\mathbf{x}) \) through supervised learning.
\begin{align}
    \boldsymbol{\theta} \triangleq \arg \min_{\boldsymbol{\theta'}} \sum_{i=1}^{N}  \mathcal{L} \left( \mathbf{\hat F}(\mathbf{x}_i; \boldsymbol{\theta'}), \mathbf{y}_i \right)
    \label{eq2}
\end{align}
where \( \boldsymbol{\theta} \) represents the parameters of the surrogate model, and \( \mathcal{L} \) is the loss of predicting $\mathbf{\hat F}(\mathbf{x}; \boldsymbol{\theta})$ when the true output is \( \mathbf{y}\) for a given input \( \mathbf{x}\).

\textbf{ABBO using offline training of the surrogate model.} 
As mentioned earlier, the surrogate model can be trained offline. In this approach, the surrogate model is first trained on a fixed $\mathcal{D}$ and then used during the backward pass of the optimization loop (see Fig.~\ref{fig:locality-aware}b and Algorithm~\ref{alg:gradpie_offline}).

\textbf{ABBO using online training of the surrogate model.} 
Another approach involves training the surrogate model in an online manner. In this method, during each iteration of the optimization process, the surrogate model updates itself with new input data (optionally incorporating \( N_s \) local samples). After updating, the surrogate model is used in the backward pass to estimate the gradients (see Fig.~\ref{fig:locality-aware}c and Algorithm~\ref{alg:gradpie_online}). 
To retrain the surrogate model effectively, a local sampling scheme is used. This scheme generates \( N_s \) samples from a normal distribution with a small standard deviation centered around the updated solution, $\mathbf{x}(t+1)$. This localized sampling strategy with simple rank selection process improves the surrogate model’s ability to accurately capture the local behavior of the objective function, enhancing optimization performance. 
Unlike the offline scheme, which requires many input-output pairs to cover the entire landscape of the black-box function, this approach focuses only on the optimization path, making it more efficient in terms of the total number of queries to the black-box function.

\section{Theoretical analysis}
The performance of the optimization scheme shown in Fig.~\ref{fig:locality-aware}b and~\ref{fig:locality-aware}c strongly depends on how accurately the gradient of the black-box system’s output is approximated with respect to the input.
A naïve approach involves sampling perturbed values around a target input and using the finite difference method to approximate its gradient. However, this method requires multiple queries to the black-box function for perturbed data points, thereby increasing the total number of input-output evaluations. Instead, we propose to improve gradient estimation performance by training a locality-aware surrogate model without increasing the total number of input-output pairs. We leverage the gradient theorem~\cite{WilliamsonTrotter2004}, which states that for a differentiable function \(\mathbf{F}\) and any curve \(\gamma_{\mathbf{x},\mathbf{x}'}\) starting at \(\mathbf{x}\) and ending at \(\mathbf{x'}\), we have
\begin{align}
    &\int_{\gamma_{\mathbf{x},\mathbf{x}'}} \nabla\mathbf{F} (\mathbf{r}) \cdot d\mathbf{r} = \mathbf{F}(\mathbf{x'})- \mathbf{F}(\mathbf{x}).
    \label{eq3}
\end{align}
To incorporate a learnable model, we replace \(\mathbf{F}(\mathbf{x})\)  inside the integral with $\mathbf{\hat F}(\mathbf{x}; \boldsymbol{\theta})$ —a surrogate model— while retaining \(\mathbf{F}\) on the right-hand side of \eqref{eq3}. Under this substitution, the equality becomes an approximation, which holds when \(\nabla\mathbf{\hat F}(\mathbf{x}; \boldsymbol{\theta}) \) closely estimates \(\nabla\mathbf{F} \). To enforce this, we therefore need to find $ \boldsymbol{\theta}$ such that the averaged difference between the LHS and RHS of \eqref{eq3} is minimized which is expressed through the loss function
\begin{multline}
    \scriptsize
    \mathcal{L}_\text{GradPIE}(\boldsymbol{\theta}) 
    = 
    E_{(\mathbf{x},\mathbf{x}')\sim\mathcal{D}^2} 
    \Biggl[\\\scriptsize
    \left\lVert \int_{\gamma_{\mathbf{x},\mathbf{x}'}} \nabla\mathbf{\hat F}(\mathbf{r}; \boldsymbol{\theta}) \cdot d\mathbf{r}
    \quad - \bigg( \mathbf{F}(\mathbf{x}')- \mathbf{F}(\mathbf{x}) \bigg)    \right\rVert_1 \Biggr],
    \label{eq:gradpie_loss_0}
\end{multline}

It measures the discrepancy between the integral of the gradient of  \(\mathbf{\hat F}(\mathbf{x}; \boldsymbol{\theta})\) over the path from \(\mathbf{x}\) to \(\mathbf{x'}\) and the difference in the target values \( \mathbf{F}(\mathbf{x}') - \mathbf{F}(\mathbf{x}) \). 
We can further simplify the above loss function into the following form by applying the gradient theorem once more to  \(\mathbf{\hat F}(\mathbf{x}; \boldsymbol{\theta})\) yielding
\begin{multline}
    \scriptsize
    \mathcal{L}_\text{GradPIE}(\boldsymbol{\theta}) 
    = E_{(\mathbf{x},\mathbf{x}')\sim\mathcal{D}^2} 
    \biggl[\\\scriptsize
    \Bigl\lVert \bigl(\mathbf{\hat F}(\mathbf{x}';\boldsymbol{\theta}) 
    - \mathbf{\hat F}(\mathbf{x};\boldsymbol{\theta})\bigr) 
    - \bigl(\mathbf{F}(\mathbf{x}') - \mathbf{F}(\mathbf{x})\bigr)
    \Bigr\rVert_1
    \biggr].
    \label{eq:gradpie_loss_1}
\end{multline}
 We will refer to this loss as \textit{GradPIE} loss. In the following, we demonstrate that it learns a superior gradient surrogate compared to the standard mean absolute error (MAE) loss
\begin{equation}
    \mathcal{L}_\text{MAE}(\boldsymbol{\theta}) = E_{\mathbf{x}\sim\mathcal{D}}\left[\left\lVert \mathbf{\hat F}(\mathbf{x};\boldsymbol{\theta}) - \mathbf{F}(\mathbf{x})\right\rVert_1\right].
    \label{eq:mae_loss}
\end{equation}
When comparing \eqref{eq:gradpie_loss_1} and \eqref{eq:mae_loss}, we observe that \textit{GradPIE} is invariant to a global offset in $\mathbf{F}(\mathbf{x})$ and $\mathbf{\hat F}(\mathbf{x}; \boldsymbol{\theta})$, such as $\mathbf{F}(\mathbf{x}) + \boldsymbol{\mu}_1$ and $\mathbf{\hat F}(\mathbf{x}; \boldsymbol{\theta}) + \boldsymbol{\mu}_2$. This is a reasonable property, as our primary goal is to ensure that the gradients are similar to improve utility for optimization. It is important to emphasize that \textit{GradPIE} achieves more than simply removing, for example, the global mean from $\mathbf{F}$ and $\mathbf{\hat F}$. As we will discuss later, \textit{GradPIE} is computed locally, ensuring that the learned surrogate $\mathbf{\hat F}$ preserves the same relative relationships among function values as observed in $\mathbf{F}$. This feature enables the learning of a better gradient surrogate. Finally, note that instead of MAE, a mean squared error (MSE) formulation could have been used, which would yield similar results as discussed in this paper.

\begin{algorithm}[t]
\scriptsize
\SetAlgoInsideSkip{2pt}
\SetAlgoSkip{2pt}
\SetInd{0.5em}{0.3em} 

\DontPrintSemicolon
\SetAlgoLined

\KwIn{%
    Initial surrogate parameters $\boldsymbol{\theta}$, 
    learning rates $\eta_1$ and $\eta_2 > 0$, 
    number of epochs $L_{\text{epochs}}$, 
    number of nearest neighbors $K$,
    black-box $\mathbf{F}(\cdot)$, 
    number of optimization steps $\tau$, 
    number of local samples $N_s$, 
    standard deviation $\sigma$, 
    number of top samples $N_{\text{best}}$, 
    convergence threshold $\epsilon$
}
\KwOut{Optimized input $\mathbf{x}^*$}

\caption{Gradient-based Black-box Optimization via Online Training of the Surrogate Model Using \textit{GradPIE}.}
\label{alg:gradpie_online}

\textbf{Initialize:}
$\boldsymbol{\theta}^{(0)} \gets \boldsymbol{\theta}$\;
Dataset $\mathcal{D}^{(0)} = \{(\mathbf{x}_i, \mathbf{F}(\mathbf{x}_i))\}_{i=1}^{N_{\text{init}}}$, where $\mathbf{x}_i \sim \mathcal{N}(0, I)$. Precompute \(\mathbf{x'}_k\) for each $\mathbf{x}_i \in \mathcal{D}^{(0)}$\;

\For{$t = 1$ \KwTo $\tau-1$}{
    \tcp*[r]{\scriptsize Generate $N_s$ new samples around $\mathbf{x}^{(t)}$ (local sampling)}
    $\mathbf{\hat{x}}_i^{(t)} \sim \mathcal{N}(\mathbf{x}^{(t)}, \sigma^2 \mathbf{I}_d), \quad i = 1, 2, \dots, N_s$\;

    \tcp*[r]{\scriptsize Update dataset with new samples}
    $\mathcal{D}^{(t+1)} \gets \mathcal{D}^{(t)} \cup \{\mathbf{\hat{x}}_i^{(t)}, \mathbf{F}(\mathbf{\hat{x}}_i^{(t)})\}$\;

    \textit{Re-train surrogate model:} \\
    \For{$l = 1$ \KwTo $L_{\text{epochs}}-1$}{
        \ForEach{mini-batch $\mathcal{B} \subseteq \mathcal{D}^{(t+1)}$}{
            Compute $\mathcal{L}_{\text{GradPIE}}$ using Eq. \eqref{eq:gradpie_loss} for $\mathcal{B}$\;
        }
        $\boldsymbol{\theta}^{(l+1)} \gets \boldsymbol{\theta}^{(l)} 
        - \eta_1 \nabla_{\boldsymbol{\theta}}\mathcal{L}_{\text{GradPIE}}(\boldsymbol{\theta})\big|_{\boldsymbol{\theta}=\boldsymbol{\theta}^{(l)}}$ 
        \tcp*[r]{\scriptsize Update surrogate parameters}
        \If{$\mathcal{L}_{\text{GradPIE}} < \epsilon$}{
            \textbf{Break} \tcp*[r]{\scriptsize Stop training if loss converges}
        }
    }
    \textit{Rank selection:} \tcp*[r]{\scriptsize Select top $N_{\text{best}}$ samples}
    Sort $\mathbf{F}(\mathbf{\hat{x}}_i^{(t)})$ in descending order and select top $N_{\text{best}}$ samples\;

    $\mathbf{x}^{(t+1)} \gets \mathbf{x}^{(t)} - \eta_2\,\nabla_{\mathbf{x}}\hat{\mathbf{F}}(\mathbf{x}; \boldsymbol{\theta})\big|_{\mathbf{x}=\mathbf{x}^{(t)}}$ 
    \tcp*[r]{\scriptsize Update $\mathbf{x}$ using surrogate gradients}
}

\textbf{Return:} optimized input $\mathbf{x}^*$.
\end{algorithm}

\begin{figure*}[t!]
  \includegraphics[width=\linewidth,trim=0 30 0 0]{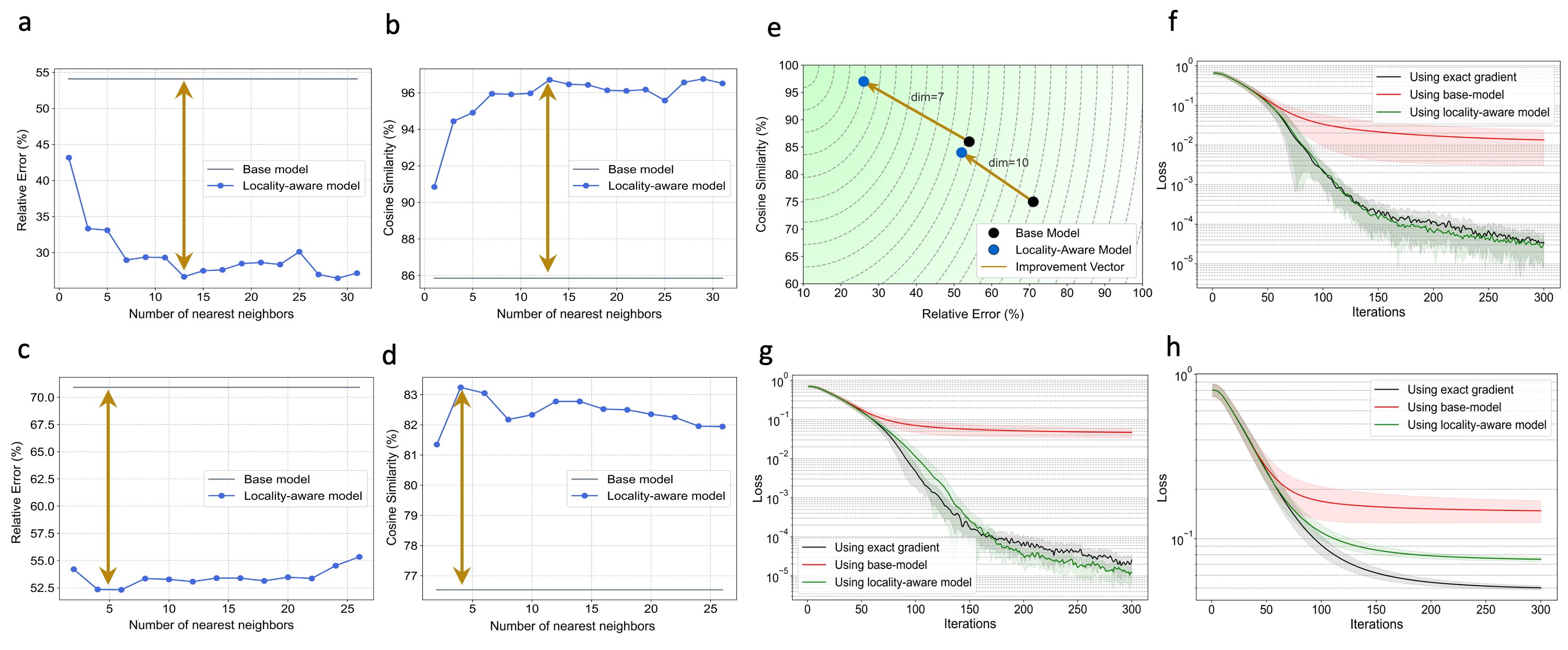}
  \caption{Relative error and cosine similarity between estimated and exact gradient for input dimensions of \( D_i = 7 \) (\textbf{a} and \textbf{b}) and \( D_i = 10 \) (\textbf{c} and \textbf{d} ), as a function of the number of nearest neighbors for CNON. \textbf{e} The improvement in gradient estimation for optimal number of nearest neighbors. Performance of ABBO using the offline-trained surrogate model for the CNON task for \textbf{f} $\mathfrak{\lambda}=0.50$, \textbf{g} $\mathfrak{\lambda}=0.55$, and \textbf{h} $\mathfrak{\lambda}=0.70$.}
  \label{fig:CNON}
\end{figure*}

We now demonstrate that the \textit{GradPIE} loss in \eqref{eq:gradpie_loss_1} minimizes the difference of the Jacobians $\mathbf{J}[\mathbf{F}]\in\R^{D_o\times D_i}$ and $\mathbf{J}[\mathbf{\hat F}]\in\R^{D_o\times D_i}$, requiring only paired samples $(\mathbf{x}_i, \mathbf{F}(\mathbf{x}_i))$ for training. By applying the gradient theorem, \eqref{eq:gradpie_loss_1} can be rewritten as
\begin{multline}
    \scriptsize
    \mathcal{L}_\text{GradPIE}(\boldsymbol{\theta}) = \\\scriptsize
    E_{(\mathbf{x},\mathbf{x}')\sim\mathcal{D}^2}\left[\left\lVert \int_{\gamma_{\mathbf{x},\mathbf{x}'}} \left(\mathbf{J}[\mathbf{\hat F}](\mathbf{u};\boldsymbol{\theta}) - \mathbf{J}[\mathbf{F}](\mathbf{u})\right) d\mathbf{u} \right\rVert_1\right],
    \label{eq:gradpie_int}
\end{multline}
where $\gamma_{\mathbf{x},\mathbf{x}'}$ is a differentiable curve connecting $\mathbf{x}$ and $\mathbf{x}'$. For simplicity, we fix $\mathbf{x}$ and assume $\lVert\mathbf{x} - \mathbf{x}'\rVert \leq \epsilon$, focusing on \eqref{eq:gradpie_int} for close samples. Under these conditions, the integral can be approximated as
\begin{multline}
    \scriptsize
   \int_{\gamma_{\mathbf{x},\mathbf{x}'}}  \left(\mathbf{J}[\mathbf{\hat F}](\mathbf{u};\boldsymbol{\theta}) - \mathbf{J}[\mathbf{F}](\mathbf{u})\right) d\mathbf{u} \approx \\\scriptsize
   \bigl(\mathbf{J}[\mathbf{\hat F}](\mathbf{x};\boldsymbol{\theta}) - \mathbf{J}[\mathbf{F}](\mathbf{x})\bigr)\bigl(\mathbf{x}' - \mathbf{x}\bigr)
\end{multline}
and this transforms \eqref{eq:gradpie_int} into
\begin{multline}
\scriptsize
\mathcal{L}_\text{GradPIE}(\boldsymbol{\theta}) \approx \\\scriptsize
\int_{\lVert\mathbf{x} - \mathbf{x}'\rVert \leq \epsilon} \left\lVert\bigl(\mathbf{J}[\mathbf{\hat F}](\mathbf{x};\boldsymbol{\theta}) - \mathbf{J}[\mathbf{F}](\mathbf{x})\bigr)\bigl(\mathbf{x}' - \mathbf{x}\bigr)\right\rVert_1 p(\mathbf{x},\mathbf{x}')d\mathbf{x}'\\\scriptsize
\approx \frac{2 \pi^{\frac{N-1}{2}}\epsilon^{N+1}}{(N+1) \Gamma\left(\frac{N-1}{2}+1\right)}p(\mathbf{x},\mathbf{x})\left\lVert\mathbf{J}[\mathbf{\hat F}](\mathbf{x};\boldsymbol{\theta}) - \mathbf{J}[\mathbf{F}](\mathbf{x})\right\rVert^\text{row}
\label{eq:integral_approx}
\end{multline}
as $\int_{\lVert\mathbf{u}\rVert \leq \epsilon}\lvert\mathbf{a}^T\mathbf{u}\rvert d\mathbf{u} = \frac{2\pi^{\frac{N-1}{2}}\epsilon^{N+1}}{(N+1)\Gamma(\frac{N-1}{2}+1)}\lVert\mathbf{a}\rVert$ with $\Gamma(.)$ denoting the gamma function and assuming that $p(\mathbf{x},\mathbf{x}')$ is continuous. Furthermore, we used $\lVert\mathbf{A}\rVert^\text{row}$ to denote the sum of the 2-norms of all rows of $\mathbf{A}$, i.e., $\lVert\mathbf{A}\rVert^\text{row} = \sum_{d=1}^{D_o}\lVert\mathbf{A}_{d:}\rVert$. Hence, from \eqref{eq:integral_approx} we can conclude that minimizing the \textit{GradPIE} loss in \eqref{eq:gradpie_loss_1} corresponds to minimizing the difference between the Jacobians of $\mathbf{F}$ and~$\mathbf{\hat F}$.

\section{Scalable Algorithm: Locality-aware surrogate model}
A naïve computation of \eqref{eq:gradpie_loss_1} requires iterating over all pairs of training inputs, which is computationally expensive. To reduce this overhead, we use the \( k \)-nearest neighbors for each sample instead. 
Given the dataset \( \mathcal{D} \), we first compute the \( k \)-nearest neighbors for each sample \( \mathbf{x} \) (see Fig.~\ref{fig:locality-aware}a) and then calculate following loss during training:
\begin{multline}
    \scriptsize
    \mathcal{L}_\text{GradPIE}(\boldsymbol{\theta}) 
    =  E_{\mathbf{x}\sim\mathcal{D}} 
    \Biggl[\\\scriptsize \frac{1}{K} \sum_{k=1}^{K}
    \Bigl\lVert \bigl(\mathbf{F}(\mathbf{x}) - \mathbf{F}(\mathbf{x'}_k)\bigr) - \bigl(\mathbf{\hat F}(\mathbf{x};\boldsymbol{\theta}) 
    - \mathbf{\hat F}(\mathbf{x'}_k;\boldsymbol{\theta})\bigr)
    \Bigr\rVert_1
    \Biggr],
    \label{eq:gradpie_loss}
\end{multline}
where \( \mathbf{x'}_k \) represents the \( k \)-th nearest neighbor of \( \mathbf{x} \). In the next section, we will empirically show how varying the number of nearest neighbors affects the accuracy of gradient estimation. We will refer to models trained with the GradPIE loss \eqref{eq:gradpie_loss} as \textit{locality-aware} models. Algorithm~\ref{alg:gradpie_online} and \ref{alg:gradpie_offline} give the algorithmic description of how we train them.

The advantage of our approach, compared to numerical gradient estimation, lies in the ability of deep surrogate models to capture more complex approximations of the objective function gradient than a linear approximation, making them particularly effective for surfaces with high curvature.
Employing such deep neural networks as surrogate models also offers further benefits, such as Hessian estimation for second-order optimization algorithms, uncertainty quantification, and the potential for automatic identification of low-dimensional parameter manifolds.

\begin{table*}[t!]
\centering
\renewcommand{\arraystretch}{0.85}
\resizebox{\textwidth}{!}{%
\begin{tabular}{lcccccc}
\toprule
\multicolumn{7}{c}{\textbf{(1) CNON Task}} \\
 & \multicolumn{3}{c}{\textbf{\( \mathbf{N_s} \) = 0}} & \multicolumn{3}{c}{\textbf{\( \mathbf{N_s} \) = 1}} \\
\cmidrule(lr){2-4} \cmidrule(lr){5-7}
\textbf{Method}
  & \textbf{50 iter} & \textbf{100 iter} & \textbf{200 iter}
  & \textbf{50 iter} & \textbf{100 iter} & \textbf{200 iter} \\
\midrule
Exact gradient
  & 0.101 $\pm$ 0.013 & 0.057 $\pm$ 0.003 & 0.048 $\pm$ 0.001
  & 0.085 $\pm$ 0.017 & 0.054 $\pm$ 0.004 & 0.048 $\pm$ 0.001 \\
\hdashline
Base-model
  & 0.261 $\pm$ 0.023 & 0.188 $\pm$ 0.044 & 0.131 $\pm$ 0.041
  & 0.194 $\pm$ 0.029 & 0.151 $\pm$ 0.026 & 0.114 $\pm$ 0.030 \\
\textbf{Locality-aware model}
  & \textbf{0.198 $\pm$ 0.025} & \textbf{0.124 $\pm$ 0.031} & \textbf{0.081 $\pm$ 0.015}
  & \textbf{0.186 $\pm$ 0.016} & \textbf{0.093 $\pm$ 0.013} & \textbf{0.066 $\pm$ 0.007} \\
\midrule
\multicolumn{7}{c}{\textbf{(2) OpAmp Task}} \\
 & \multicolumn{3}{c}{\textbf{Mean} $\pm$ \textbf{Std}} & \multicolumn{3}{c}{\textbf{IQR:} Median (25--75\%)} \\
\cmidrule(lr){2-4} \cmidrule(lr){5-7}
\textbf{Method}
  & \textbf{50 iter} & \textbf{100 iter} & \textbf{200 iter}
  & \textbf{50 iter} & \textbf{100 iter} & \textbf{200 iter} \\
\midrule
Random
  & 7.074 $\pm$ 0.904 & 7.330 $\pm$ 0.849 & 7.628 $\pm$ 0.835
  & 7.036 (6.384--7.701) & 7.330 (6.714--7.902) & 7.610 (7.032--8.224) \\
Base-model
  & 8.419 $\pm$ 1.010 & 8.628 $\pm$ 0.947 & 8.860 $\pm$ 0.899
  & 8.460 (7.785--9.096) & 8.689 (8.058--9.311) & 8.966 (8.329--9.499) \\
\midrule
\textbf{Locality-aware model}
  & \textbf{9.011 $\pm$ 0.828} & \textbf{9.389 $\pm$ 0.727} & \textbf{9.615 $\pm$ 0.694}
  & \textbf{9.068 (8.582--9.473)} & \textbf{9.430 (9.053--9.763)} & \textbf{9.700 (9.292--9.952)} \\
\midrule
\multicolumn{7}{c}{\textbf{(3) OWMS Task}} \\
 & \multicolumn{3}{c}{\textbf{Mean} $\pm$ \textbf{Std}} & \multicolumn{3}{c}{\textbf{IQR:} Median (25--75\%)} \\
\cmidrule(lr){2-4} \cmidrule(lr){5-7}
\textbf{Method}
  & \textbf{50 iter} & \textbf{100 iter} & \textbf{200 iter}
  & \textbf{50 iter} & \textbf{100 iter} & \textbf{200 iter} \\
\midrule
Random
  & 0.239 $\pm$ 0.012 & 0.233 $\pm$ 0.009 & 0.227 $\pm$ 0.009
  & 0.236 (0.232--0.249) & 0.235 (0.229--0.238) & 0.231 (0.221--0.235) \\
BayOpt
  & 0.147 $\pm$ 0.023 & 0.120 $\pm$ 0.020 & 0.103 $\pm$ 0.019
  & 0.142 (0.133--0.155) & 0.119 (0.109--0.129) & 0.104 (0.093--0.114) \\
Base-model
  & 0.172 $\pm$ 0.018 & 0.140 $\pm$ 0.015 & 0.117 $\pm$ 0.013
  & 0.172 (0.159--0.186) & 0.142 (0.132--0.150) & 0.118 (0.109--0.125) \\
\midrule
\textbf{Locality-aware model}
  & \textbf{0.133 $\pm$ 0.014} & \textbf{0.107 $\pm$ 0.016} & \textbf{0.090 $\pm$ 0.015}
  & \textbf{0.132 (0.123--0.142)} & \textbf{0.105 (0.095--0.118)} & \textbf{0.087 (0.078--0.102)} \\
\bottomrule
\end{tabular}%
}
\vspace{-0.2cm}
\caption{Performance of ABBO using the online-trained surrogate model across three tasks: CNON, OpAmp, and OWMS. For the CNON task, results are shown for {\( {N_s} \) = 0} and {\( {N_s} \) = 1}. For the OpAmp and OWMS tasks, we report both the mean ± standard deviation and the median with interquartile range (IQR) at 50, 100, and 200 iterations. Note that smaller values indicate better performance for CNON and OWMS, while larger values indicate better performance for the OpAmp task.}
\label{table:all_task}
\end{table*}
\section{Experiments}
This section outlines the tasks used to compare the performance of the proposed locality-aware surrogate models with baseline models. Our empirical studies were conducted on three tasks from different engineering domains:
\begin{enumerate}[label=\textbf{\arabic*.},left=0pt,itemsep=0pt,topsep=0pt,partopsep=0pt,parsep=0pt]
    \item Coupled Nonlinear Oscillator Network (CNON): We used Coupled Nonlinear Oscillators —dynamic systems defined by nonlinear interactions between oscillatory components \cite{wright2022deep,lanthaler2024neural}— to assess gradient estimation performance compared to the exact gradient. ABBO optimization using offline and online training of the surrogate model experiments was conducted to benchmark results against the ideal scenario with direct access to the exact gradient.
    
    \item Analog Integrated Circuits (operational amplifier: OpAmp): The second task involves optimizing a two-stage operational amplifier (OpAmp) circuit by maximizing a figure-of-merit (FOM), which is defined as a combination of open-loop gain, unity-gain bandwidth, and phase margin.
    
    \item Optical Wave Manipulation System (OWMS): The final task involves spatial wavefront shaping through the optimization of Spatial Light Modulator (SLM) phase patterns in an optical system \cite{filipovich2024torchoptics}. The optimization is defined over a 3600-dimensional continuous search space.
\end{enumerate}

\subsection{CNON}
To empirically evaluate the effectiveness of the proposed \textit{GradPIE} loss function, we start with a toy example: a Coupled Nonlinear Oscillator Network (CNON) described by the following equations of motion
\begin{equation}
    \small
    \frac{d^2 \mathfrak{q}_i}{dt^2} = -\sin(\pi \mathfrak{q}_i) 
    + \sum_{j=1}^N \mathfrak{J}_{ij} \big(\sin(\pi \mathfrak{q}_j) - \sin(\pi \mathfrak{q}_i)\big) 
    + \mathfrak{e}_i,
    \label{eq:CNO}
\end{equation}
where $\mathfrak{q}_i$ are the oscillator amplitudes, \(\mathfrak{J}_{ij}\) are the symmetric coupling coefficients, and $\mathfrak{e}_i$ are the individual oscillator drives. Eq.~\eqref{eq:CNO} is often used as approximation to represent the Frenkel-Kontorova model, a widely used model in condensed matter physics \cite{braun1998nonlinear, wright2022deep}. 

We encode the input data as the initial amplitudes, $\mathfrak{q}_i^{0}$, and take the output to be the state of the oscillator after some time evolution, $\mathfrak{q}_i^{\infty}$ (see Appendix for details).

To evaluate the effectiveness of gradient estimation, we train a multilayer neural network using two loss functions: $\mathcal{L}_\text{MAE}$ and $\mathcal{L}_\text{GradPIE}$, with a dataset containing $N = 1000$ input-output pairs. We then assess the performance of gradient estimation by calculating the relative error, $\text{Relative Error} = \frac{\|\mathfrak{g}_\text{est} - \mathfrak{g}_\text{exact}\|}{\|\mathfrak{g}_\text{exact}\|}$, and the cosine similarity, $\cos(\theta) = \frac{\mathfrak{g}_\text{est} \cdot \mathfrak{g}_\text{exact}}{\|\mathfrak{g}_\text{est}\| \|\mathfrak{g}_\text{exact}\|}$, between the estimated gradient vector, $\mathfrak{g}_\text{est}$, and the exact gradient vector, $\mathfrak{g}_\text{exact}$.

The relative error and cosine similarity for dimensions $D_i = 7$ and $D_i = 10$ as a function of the number of nearest neighbors are shown in Figs.~\ref{fig:CNON}a and b and Figs.~\ref{fig:CNON}c and d, respectively. As illustrated, the \textit{GradPIE} loss significantly improves gradient estimation performance. This improvement is reflected in relative error reductions of over 27\% and 20\%, and cosine similarity increases of more than 10\% and 8\%, respectively. In this experiment, increasing the number of nearest neighbors does not necessarily lead to improved performance. The optimal value of this parameter is highly correlated with the dimensionality and complexity of the problem, making it a case-dependent hyperparameter.
In Fig.~\ref{fig:CNON}e, we summarize the improvement in gradient estimation in terms of relative error (x-axis) and cosine similarity (y-axis) for the optimal number of nearest neighbors.

We perform ABBO optimization using offline training of the surrogate model for a CNON with a dimensionality of $D_i = 7$. The optimization problem is defined as
\begin{align}
    \boldsymbol{{\mathfrak{q}^{0}}^*} \triangleq \arg \min_{\boldsymbol{\mathfrak{q}^{0}}}  \| \mathfrak{q}^{\infty} - \boldsymbol{\lambda} \|_1,
    \label{eq4}
\end{align}
where $\mathfrak{q}^{0}$ and $\mathfrak{q}^{\infty}$ represent the initial and final states of the oscillator amplitude vectors, respectively. $\mathfrak{q}^{0*}$ denotes the optimal initial state of the oscillator amplitude vector, and $\boldsymbol{\lambda}$ is a fixed target vector. The goal is to minimize the $\ell_{1\text{-norm}}$ of the deviation between $\mathfrak{q}^{\infty}$ and $\boldsymbol{\lambda}$, ensuring the system converges to the desired configuration. The surrogate model is initially trained on a dataset consisting of $N = 1000$ input-output pairs. Once trained, the model is used within a gradient-based optimization scheme (illustrated in Fig.~\ref{fig:locality-aware}b) to evaluate its performance across three different cases: the base model, the locality-aware model, and a baseline case using the exact gradient for comparison. Figures~\ref{fig:CNON}f–h present the results for three distinct values of $\boldsymbol{\lambda}$. The plots clearly show the superior performance of the locality-aware model compared to the base model, demonstrating the effectiveness of incorporating the \textit{GradPIE} loss function in improving optimization performance.


An ABBO optimization using online training of the surrogate model is also implemented for a CNON with a dimensionality of $D_i = 10$. In this framework, the surrogate model is iteratively updated during the gradient-based optimization loop using newly updated input-output pairs. Once updated, the surrogate model is frozen and used to estimate gradients in subsequent iterations. The results for {\( {N_s} \) = 0} and {\( {N_s} \) = 1}, covering three scenarios - using the exact gradient, the baseline model, and the locality-aware model (with the \textit{GradPIE} loss function) - are shown in Figs.~\ref{fig:OPAMP}a and b, respectively. A quantitative comparison of the different methods’ performance at 50, 100, and 200 iterations is provided in Table~\ref{table:all_task}.
In both cases, the locality-aware model outperforms the base model and closely approaches the performance achieved using the exact gradient. For {\( {N_s} \) = 1}, the optimization performance aligns more closely with the exact gradient case, especially after the 100th iteration. This improvement is attributed to the inclusion of the local sampling scheme.

\subsection{OpAmp}
\begin{figure}[t!]
    \includegraphics[width=\linewidth,trim=0 23 0 0]{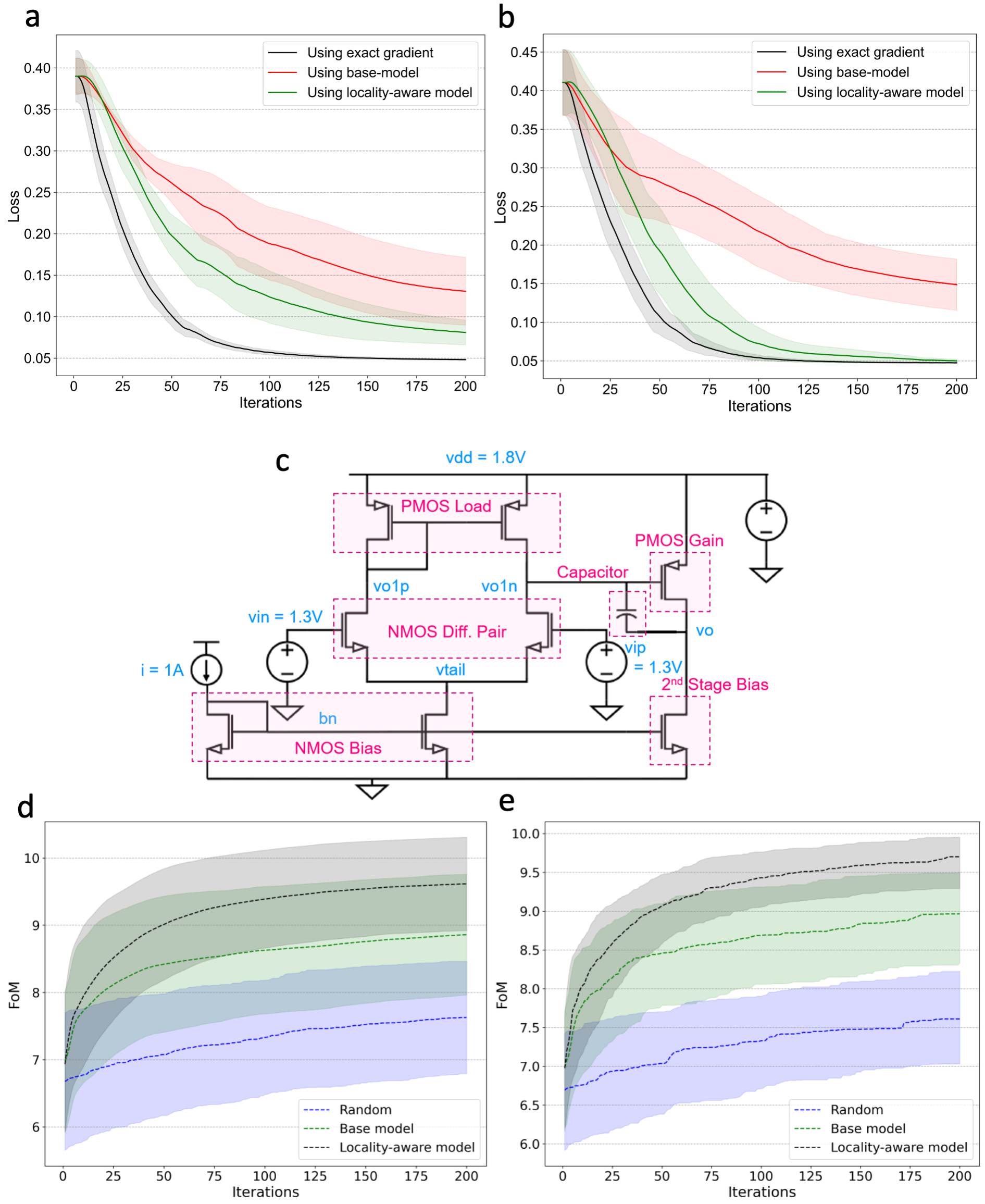}
    \caption{Performance of ABBO using the online-trained surrogate model for the CNON task (\( \boldsymbol{\lambda} = [-0.55, 0.125, 0.31, -0.38, 0.60] \)): \textbf{a} without local sampling (\( N_s = 0 \)) and \textbf{b} with local sampling (\( N_s = 1 \)). \textbf{c} Schematic of the OpAmp circuit task. \textbf{d} Mean ± standard deviation and \textbf{e} median with interquartile range (IQR) of ABBO using the online-trained surrogate model for the OpAmp task.}
    \label{fig:OPAMP}
\end{figure}

As a second task, we explore the use of ABBO optimization for sizing a two-stage OpAmp circuit, as shown in Fig.~\ref{fig:OPAMP}c. Specifically, the problem involves sizing the length and width of the MOSFET transistors and the value of a capacitor, resulting in an optimization problem with $D=10$ variables. The goal is to maximize the figure of merit (FoM):
\begin{equation}
    \text{FoM} = 1.2\cdot\frac{\text{G}}{100} + 1.6\cdot\frac{\text{PM}}{90^\circ} + 10\cdot\frac{\text{UGBW}}{1 \text{ GHz}}
\end{equation}
where $\text{G}$ denotes the linear DC gain, $\text{PM}$ the phase margin in degrees, and $\text{UGBW}$ the unity-gain bandwidth in Hz. This FoM, proposed in \cite{dong2023cktgnn}, balances multiple objectives for an OpAmp design.

Using an online approach, we aim to maximize the FoM. Since the performance of gradient ascent depends on the initial starting point, we first draw $500$ random sizings uniformly from the allowed sizing ranges. In each iteration, we update the surrogate model by performing $100$ epochs of supervised training on the available data. Then, we apply one step of gradient ascent using Adam on the $5$ best samples, with gradients provided by the surrogate model. Overall, we do this for a total of $200$ iterations. 

Figs.~\ref{fig:OPAMP}d and e present the results for 200 randomly chosen initial points, where we also compare against a random baseline that selects $5$ samples randomly within the allowed sizing ranges. Comparing the base model with the proposed locality-aware model, we observe that our method performs significantly better. This improvement is also evident in the results shown in Table~\ref{table:all_task}. For example, using the gradient from our locality-aware model with $50$ iterations yields better results than using the gradient from the base model with $200$ iterations, indicating that we require far fewer active samples to be collected.

\begin{figure*}[t!]
\includegraphics[width=\linewidth,trim=0 20 0 0]{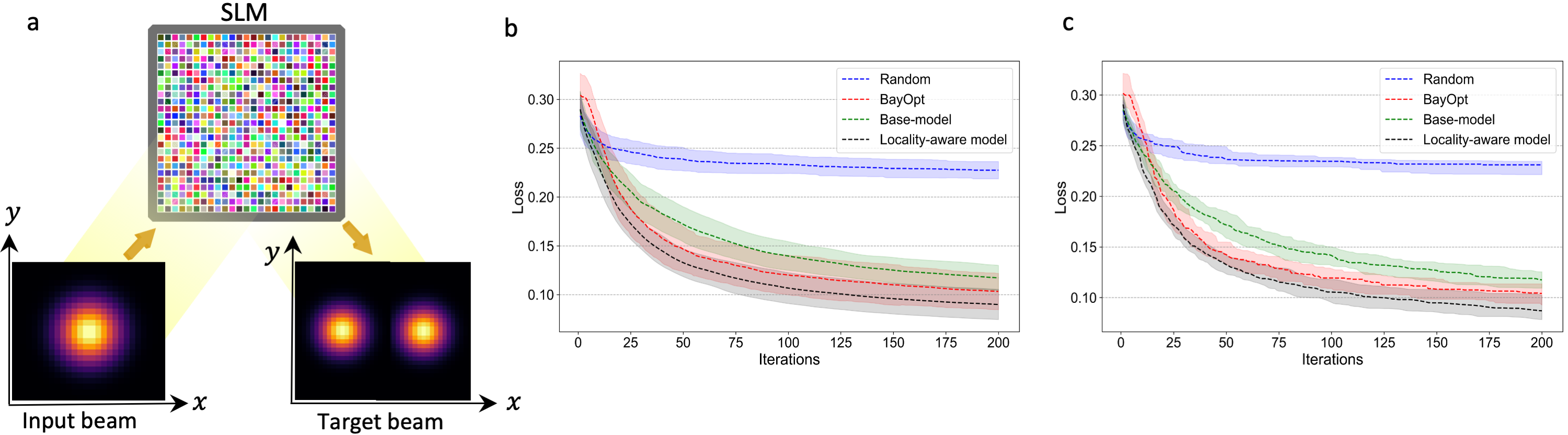}
\caption{\textbf{a} Schematic
of the OWMS. An input Gaussian beam is illuminated onto a Spatial Light Modulator (SLM) with 3600 parameters to be optimized such that the output waveform matches the target waveform. \textbf{b} Mean ± standard deviation and \textbf{c} median with interquartile range (IQR) of ABBO using the online-trained surrogate model for the OWMS task.}
\label{fig:OWMS}
\end{figure*}
\subsection{OWMS}
Optical wave manipulation has numerous practical applications, ranging from imaging to the emerging field of optical neural networks, where light in optical setups is used for performing neural computation \cite{momeni2023backpropagation, momeni2023phyff}. A key component of such systems is the manipulation of optical waves using Spatial Light Modulation (SLM). The modulation of an input optical field, \(\mathfrak{\psi}_{\text{in}}(x, y)\), by a complex-valued modulation profile \(\mathfrak{M}\), is achieved as a elementwise product. The resulting output field \(\mathfrak{\psi}_{\text{out}}(x, y)\), is expressed as
\begin{equation}
    \mathfrak{\psi}_{\text{out}}(x, y) = \mathfrak{M} \cdot \mathfrak{\psi}_{\text{in}}(x, y).
  \end{equation}
The free-space propagation of optical fields between planes is modeled using scalar diffraction theory. According to the Rayleigh--Sommerfeld diffraction integral, the propagation of an input field \(\mathfrak{\psi}_0(x', y')\) at \(z = 0\) to a position \((x, y, z)\) is given by
\begin{equation}
    \mathfrak{\psi}_z(x, y) = \iint \mathfrak{h}_z(x - x', y - y') \mathfrak{\psi}_0(x', y') \, dx' \, dy'.
    \label{optics_integral}
\end{equation}
Here, the impulse response \(\mathfrak{h}_z\) is expressed as
\begin{equation}
    \mathfrak{h}_z(x - x', y - y') = \frac{z}{i \mathfrak{\lambda} \mathfrak{d}} \left( 1 + \frac{i}{\mathfrak{k} \mathfrak{d}} \right) \exp(i \mathfrak{k} \mathfrak{d}),
\end{equation}
where \(\mathfrak{d} = \sqrt{(x - x')^2 + (y - y')^2 + z^2}\) is the Euclidean distance between the input and output positions, \(\mathfrak{\lambda}\) is the wavelength, and \(\mathfrak{k} = \frac{2\pi}{\mathfrak{\lambda}}\) is the wavenumber.
We use TorchOptics \cite{filipovich2024torchoptics} to numerically evaluate the diffraction integral in \eqref{optics_integral} using the fast Fourier transform (FFT) by treating it as a convolution.

We consider a Gaussian beam as the input field, with a waist radius of 70 $\mu$m and a wavelength of 700 nm. The objective of the optimization is to determine the complex values of $\mathfrak{M}$ such that the input Gaussian beam splits into two Gaussian beams at the output (see Fig.~\ref{fig:OWMS}a). The dimension of $\mathfrak{M}$ is $60 \times 60$, resulting in a total of 3600 parameters to be optimized. We perform ABBO optimization by leveraging the fact that all computations for calculating the optical field and the objective function—defined as the difference in the $\ell_1$-norm between the target field and the output field, $\|\mathfrak{\psi}_{\text{out}}(x, y) - \mathfrak{\psi}_{\text{target}}(x, y)\|_1$—are conducted through a black-box simulator implemented using TorchOptics. We also implement Random and BayOpt schemes under the same settings as those used for the base model and the locality-aware model, all evaluated over 100 randomly chosen initial points. The corresponding results, including the mean $\pm$ standard deviation (std) and interquartile range (IQR) of the loss values, are presented in Figs.~\ref{fig:OWMS}b and c, as well as in Table~\ref{table:all_task}. The results show that the locality-aware model outperforms all the aforementioned approaches.

\begin{table}[t!]
\centering
\footnotesize
\resizebox{\columnwidth}{!}{%
  \begin{tabular}{lccc}
    \toprule
      \multicolumn{4}{c}{\textbf{(1) CNON Task}} \\
    & \multicolumn{2}{c}{
       \begin{tabular}[c]{@{}c@{}}
         \textbf{Budget} $O(\cdot)$ \\ 
         \text{to reach \textbf{Base-model} performance}
       \end{tabular}
     }
    & \quad\textbf{Reduction Ratio}\\
    \cmidrule(lr){2-3}
    \textbf{Method} &\qquad \( \mathbf{N_s = 0} \) & \quad \( \mathbf{N_s = 1} \) & \\
    \midrule
    Locality-aware model& \qquad $O(100)$ & \quad $O(120)$ & \(\sim50\% (N_s = 0)\) and \(\sim70\% (N_s = 1)\) \\
    \bottomrule
    \end{tabular}
}

\vspace{1em} 

\resizebox{\columnwidth}{!}{%
  \begin{tabular}{lcc}
  \toprule
  \multicolumn{3}{c}{\textbf{(2) OpAmp Task}} \\
  \textbf{Method} 
  & \multicolumn{1}{c}{ 
       \begin{tabular}[c]{@{}c@{}}
         \textbf{Budget} $O(\cdot)$ \\ 
         \text{to reach \textbf{Base-model} performance}
       \end{tabular}
     }
  & \textbf{Reduction Ratio} \\
  \midrule
  Locality-aware model & \(O(1000)\) & \(\sim60\%\) \\
  \bottomrule
  \end{tabular}%

}

\vspace{1em} 

\resizebox{\columnwidth}{!}{%
  \begin{tabular}{lcc}
  \toprule
  \multicolumn{3}{c}{\textbf{(3) OWMS Task}} \\
  \textbf{Method} 
  & \multicolumn{1}{c}{
       \begin{tabular}[c]{@{}c@{}}
         \textbf{Budget} $O(\cdot)$ to reach\\ 
         \text{\textbf{Base-model/BayOpt} performance}
       \end{tabular}
     }
  & \textbf{Reduction Ratio} \\
  \midrule
  Locality-aware model & \(O(750)/O(1200)\) & \([\sim62\%/\sim40\%]\) \\
  \bottomrule
  \end{tabular}%
  }
\vspace{-0.3cm}
\caption{Budget \(O(.)\) of black-box calls required for locality-aware models to match the performance of the Base-model or BayOpt after 200 iterations.}
\label{table:budget_comparison}
\end{table}

 \subsection{Discussion of the Results}
We also implement another analysis on the number of black-box queries required by locality-aware models to match the performance of the base model or BayOpt after 200 iterations across all tasks (see Table \ref{table:budget_comparison}). Notably, the reduction ratio exceeds \(40\%\) for all tasks, highlighting the effectiveness of our approach.
From the results, we can conclude that using our proposed \textit{GradPIE} loss allows to learn surrogate models with enhanced gradient estimation accuracy, which is critical for optimization tasks. These models are particularly effective in complex, high-dimensional systems (e.g., CNON and OWMS), where exact gradients are unavailable or computationally expensive to compute.
There are some important points that need to be highlighted here.

First, as mentioned in the Appendix, simple multilayer perceptrons (MLPs) are used as surrogate models for learning gradients. However, other surrogate models that can be trained using \textit{GradPIE} could also be employed to improve gradient estimation performance. For instance, generative adversarial networks (GANs) can be used for the same purpose \cite{shirobokov2020black}.

Second, in Algorithm \ref{alg:gradpie_online}, we employ a simple local sampling method, which involves sampling from a normal distribution around the newly updated input with a small standard deviation. However, more advanced methods could be utilized to generate local samples. For example, the Latin Hypercube Sampling (LHS) algorithm \cite{iman1980latin}, which is well-suited for high-dimensional spaces, or approaches that leverage local gradient information to guide the sampling. 

Third, the effectiveness of utilizing surrogate models to estimate gradients for gradient-based black-box optimization, compared to other methods such as numerical differentiation with gradient descent, guided evolutionary strategies, and BayOpt, has been demonstrated in previous studies \cite{grathwohl2017backpropagation,shirobokov2020black,ruiz2018learning,louppe2019adversarial}.
 Although we compared our performance with BayOpt and random search, a fair and rigorous comparison to verify the effectiveness of the proposed locality-aware surrogate model using the \textit{GradPIE} loss function is to compare it with the exact same surrogate model trained with MSE or MAE loss functions, as demonstrated across all tasks.

\section{Related Work}
Several methods for black-box optimization exist, mainly differing in whether gradients of the objective function are available.
In non-gradient-based optimization such as those modeled by Monte Carlo simulators, only data samples from an intractable likelihood can be generated. 
Common approaches in these cases include genetic algorithms \cite{banzhaf1998genetic}, BayOpt \cite{snoek2012practical,eriksson2019scalable}, and numerical differentiation \cite{svanberg1987method}. 
However, these methods often lack scalability for high-dimensional black-box systems. Black-box gradient-based optimization methods can be broadly categorized into two classes: offline black-box optimization and active black-box optimization, depending on the system under optimization.
In scenarios where actively querying the black-box function is impossible or challenging, offline black-box optimization is more suitable. 
Conversely, active black-box optimization is well-suited for scenarios compatible with automated \textit{in silico} or \textit{in situ} experimental settings, such as simulators, robotics, smart sensors, and the training of analog neural networks.
\paragraph{Offline black-box optimization} 
A fundamental challenge in offline optimization lies in the mismatch between surrogate predictions and true objective values, particularly when extrapolating beyond training data. Conservative Objective Models (COMs) \cite{trabucco2021conservative} address this by penalizing overestimations of out-of-distribution inputs through adversarial training. Building on this, \cite{fu2021offline} reduces prediction uncertainty via normalized data likelihood maximization. \cite{yu2021roma} adopts techniques in model pre-training and adaptation to enforce a criteria of local smoothness.
Alternative strategies focus on the search process. They represent the search model as a distribution conditioned on the rare event of achieving high oracle performance
or an adaptive gradient update policy with learnable parameters \cite{brookes2019conditioning,fannjiang2020autofocused,chemingui2024offline}.
\paragraph{Active Black-Box Optimization} 
Traditional ABBO methods rely on global surrogate models with BayOpt \cite{snoek2012practical} or stochastic gradient estimators \cite{wang2018stochastic} being predominant choices. While differentiable surrogates \cite{kumar2020model,eriksson2019scalable} enable direct gradient-based optimization, they struggle with gradient misalignment in data-scarce regimes. Finite-difference approximations \cite{spall2022hybrid} provide alternative gradient estimates but become computationally prohibitive in high-dimensional spaces.
The L-GSO framework \cite{shirobokov2020black} trains deep generative models to iteratively approximate the simulator in local neighborhoods of the parameter space demonstrating improved sample efficiency for high-dimensional problems. 
Under limited query budgets, such methods often struggle to align the surrogate’s gradients with the exact gradients.

\section{Conclusion}
This study presents a novel theoretical framework for active gradient-based black-box optimization, focusing on enhancing gradient estimation through the use of locality-aware surrogate models.
Central to this framework is the proposed \textit{GradPIE} loss function, which significantly improves gradient estimation for both offline and online training of surrogate models. 
Building on this theoretical foundation, we developed a novel algorithm for creating surrogate models exhibiting good gradient matching and by this achieving superior performance across diverse real-world benchmarks. 
While the proposed theory and algorithm are rooted in the domain of active gradient-based optimization, the underlying principles have broader applicability to related fields such as reinforcement learning.

\section*{Impact Statement}
This paper introduces a novel theoretical perspective for understanding and analyzing gradient-based optimization problems through surrogate models, offering a cost-effective alternative to traditional gradient estimation via surrogate models. 
The methodological advancements and insights presented in this work have the potential to drive improvements across various science and engineering domains, well-suited for automated \textit{in silico} or \textit{in situ} experimental settings.
In particular, the training of physical neural networks—analog neural networks such as memristor-based \cite{yao2020fully, aguirre2024hardware} and optical neural networks \cite{ma2025quantum,bernstein2023single}—has recently garnered significant attention due to their exceptional energy efficiency \cite{mcmahon2023physics,momeni2023backpropagation}. One of the most widely adopted approaches for training such networks involves the use of surrogate models to estimate gradients. We posit that the proposed algorithm can substantially enhance the training of such analog neural networks, enabling more efficient and effective training schemes.

\bibliography{example_paper}

\begin{thebibliography}{47}
\providecommand{\natexlab}[1]{#1}
\providecommand{\url}[1]{\texttt{#1}}
\expandafter\ifx\csname urlstyle\endcsname\relax
  \providecommand{\doi}[1]{doi: #1}\else
  \providecommand{\doi}{doi: \begingroup \urlstyle{rm}\Url}\fi

\bibitem[Aguirre et~al.(2024)Aguirre, Sebastian, Le~Gallo, Song, Wang, Yang, Lu, Chang, Ielmini, Yang, et~al.]{aguirre2024hardware}
Aguirre, F., Sebastian, A., Le~Gallo, M., Song, W., Wang, T., Yang, J.~J., Lu, W., Chang, M.-F., Ielmini, D., Yang, Y., et~al.
\newblock Hardware implementation of memristor-based artificial neural networks.
\newblock \emph{Nature communications}, 15\penalty0 (1):\penalty0 1974, 2024.

\bibitem[Ashby(2000)]{ashby2000multi}
Ashby, M.
\newblock Multi-objective optimization in material design and selection.
\newblock \emph{Acta materialia}, 48\penalty0 (1):\penalty0 359--369, 2000.

\bibitem[Banzhaf et~al.(1998)Banzhaf, Nordin, Keller, and Francone]{banzhaf1998genetic}
Banzhaf, W., Nordin, P., Keller, R.~E., and Francone, F.~D.
\newblock \emph{Genetic programming: an introduction: on the automatic evolution of computer programs and its applications}.
\newblock Morgan Kaufmann Publishers Inc., 1998.

\bibitem[Bernstein et~al.(2023)Bernstein, Sludds, Panuski, Trajtenberg-Mills, Hamerly, and Englund]{bernstein2023single}
Bernstein, L., Sludds, A., Panuski, C., Trajtenberg-Mills, S., Hamerly, R., and Englund, D.
\newblock Single-shot optical neural network.
\newblock \emph{Science Advances}, 9\penalty0 (25):\penalty0 eadg7904, 2023.

\bibitem[Braun \& Kivshar(1998)Braun and Kivshar]{braun1998nonlinear}
Braun, O.~M. and Kivshar, Y.~S.
\newblock Nonlinear dynamics of the frenkel--kontorova model.
\newblock \emph{Physics Reports}, 306\penalty0 (1-2):\penalty0 1--108, 1998.

\bibitem[Brookes et~al.(2019)Brookes, Park, and Listgarten]{brookes2019conditioning}
Brookes, D., Park, H., and Listgarten, J.
\newblock Conditioning by adaptive sampling for robust design.
\newblock In \emph{International conference on machine learning}, pp.\  773--782. PMLR, 2019.

\bibitem[Chemingui et~al.(2024)Chemingui, Deshwal, Hoang, and Doppa]{chemingui2024offline}
Chemingui, Y., Deshwal, A., Hoang, T.~N., and Doppa, J.~R.
\newblock Offline model-based optimization via policy-guided gradient search.
\newblock In \emph{Proceedings of the AAAI Conference on Artificial Intelligence}, volume~38, pp.\  11230--11239, 2024.

\bibitem[Dao et~al.(2024)Dao, Le~Nguyen, Truong, and Hoang]{dao2024boosting}
Dao, M.~C., Le~Nguyen, P., Truong, T.~N., and Hoang, T.~N.
\newblock Boosting offline optimizers with surrogate sensitivity.
\newblock In \emph{Forty-first International Conference on Machine Learning}, 2024.

\bibitem[de~Avila Belbute-Peres et~al.(2018)de~Avila Belbute-Peres, Smith, Allen, Tenenbaum, and Kolter]{de2018end}
de~Avila Belbute-Peres, F., Smith, K., Allen, K., Tenenbaum, J., and Kolter, J.~Z.
\newblock End-to-end differentiable physics for learning and control.
\newblock \emph{Advances in neural information processing systems}, 31, 2018.

\bibitem[Degrave et~al.(2019)Degrave, Hermans, Dambre, and Wyffels]{degrave2019differentiable}
Degrave, J., Hermans, M., Dambre, J., and Wyffels, F.
\newblock A differentiable physics engine for deep learning in robotics.
\newblock \emph{Frontiers in neurorobotics}, 13:\penalty0 6, 2019.

\bibitem[Dong et~al.(2023)Dong, Cao, Zhang, Tao, Chen, and Zhang]{dong2023cktgnn}
Dong, Z., Cao, W., Zhang, M., Tao, D., Chen, Y., and Zhang, X.
\newblock Cktgnn: Circuit graph neural network for electronic design automation.
\newblock \emph{arXiv preprint arXiv:2308.16406}, 2023.

\bibitem[Eriksson et~al.(2019)Eriksson, Pearce, Gardner, Turner, and Poloczek]{eriksson2019scalable}
Eriksson, D., Pearce, M., Gardner, J., Turner, R.~D., and Poloczek, M.
\newblock Scalable global optimization via local bayesian optimization.
\newblock \emph{Advances in neural information processing systems}, 32, 2019.

\bibitem[Fannjiang \& Listgarten(2020)Fannjiang and Listgarten]{fannjiang2020autofocused}
Fannjiang, C. and Listgarten, J.
\newblock Autofocused oracles for model-based design.
\newblock \emph{Advances in Neural Information Processing Systems}, 33:\penalty0 12945--12956, 2020.

\bibitem[Filipovich \& Lvovsky(2024)Filipovich and Lvovsky]{filipovich2024torchoptics}
Filipovich, M.~J. and Lvovsky, A.
\newblock Torchoptics: An open-source python library for differentiable fourier optics simulations.
\newblock \emph{arXiv preprint arXiv:2411.18591}, 2024.

\bibitem[Fu \& Levine(2021)Fu and Levine]{fu2021offline}
Fu, J. and Levine, S.
\newblock Offline model-based optimization via normalized maximum likelihood estimation.
\newblock \emph{arXiv preprint arXiv:2102.07970}, 2021.

\bibitem[Grathwohl et~al.(2017)Grathwohl, Choi, Wu, Roeder, and Duvenaud]{grathwohl2017backpropagation}
Grathwohl, W., Choi, D., Wu, Y., Roeder, G., and Duvenaud, D.
\newblock Backpropagation through the void: Optimizing control variates for black-box gradient estimation.
\newblock \emph{arXiv preprint arXiv:1711.00123}, 2017.

\bibitem[Hutter et~al.(2011)Hutter, Hoos, and Leyton-Brown]{hutter2011sequential}
Hutter, F., Hoos, H.~H., and Leyton-Brown, K.
\newblock Sequential model-based optimization for general algorithm configuration.
\newblock In \emph{Learning and Intelligent Optimization: 5th International Conference, LION 5, Rome, Italy, January 17-21, 2011. Selected Papers 5}, pp.\  507--523. Springer, 2011.

\bibitem[Iman et~al.(1980)Iman, Davenport, and Zeigler]{iman1980latin}
Iman, R.~L., Davenport, J.~M., and Zeigler, D.~K.
\newblock Latin hypercube sampling (program user's guide).[lhc, in fortran].
\newblock Technical report, Sandia Labs., Albuquerque, NM (USA), 1980.

\bibitem[Krishnamoorthy et~al.(2022)Krishnamoorthy, Mashkaria, and Grover]{krishnamoorthy2022generative}
Krishnamoorthy, S., Mashkaria, S.~M., and Grover, A.
\newblock Generative pretraining for black-box optimization.
\newblock \emph{arXiv preprint arXiv:2206.10786}, 2022.

\bibitem[Kumar \& Levine(2020)Kumar and Levine]{kumar2020model}
Kumar, A. and Levine, S.
\newblock Model inversion networks for model-based optimization.
\newblock \emph{Advances in neural information processing systems}, 33:\penalty0 5126--5137, 2020.

\bibitem[Lanthaler et~al.(2024)Lanthaler, Rusch, and Mishra]{lanthaler2024neural}
Lanthaler, S., Rusch, T.~K., and Mishra, S.
\newblock Neural oscillators are universal.
\newblock \emph{Advances in Neural Information Processing Systems}, 36, 2024.

\bibitem[Louppe et~al.(2019)Louppe, Hermans, and Cranmer]{louppe2019adversarial}
Louppe, G., Hermans, J., and Cranmer, K.
\newblock Adversarial variational optimization of non-differentiable simulators.
\newblock In \emph{The 22nd International Conference on Artificial Intelligence and Statistics}, pp.\  1438--1447. PMLR, 2019.

\bibitem[Ma et~al.(2025)Ma, Wang, Laydevant, Wright, and McMahon]{ma2025quantum}
Ma, S.-Y., Wang, T., Laydevant, J., Wright, L.~G., and McMahon, P.~L.
\newblock Quantum-limited stochastic optical neural networks operating at a few quanta per activation.
\newblock \emph{Nature Communications}, 16\penalty0 (1):\penalty0 359, 2025.

\bibitem[McMahon(2023)]{mcmahon2023physics}
McMahon, P.~L.
\newblock The physics of optical computing.
\newblock \emph{Nature Reviews Physics}, 5\penalty0 (12):\penalty0 717--734, 2023.

\bibitem[Mennel et~al.(2020)Mennel, Symonowicz, Wachter, Polyushkin, Molina-Mendoza, and Mueller]{mennel2020ultrafast}
Mennel, L., Symonowicz, J., Wachter, S., Polyushkin, D.~K., Molina-Mendoza, A.~J., and Mueller, T.
\newblock Ultrafast machine vision with 2d material neural network image sensors.
\newblock \emph{Nature}, 579\penalty0 (7797):\penalty0 62--66, 2020.

\bibitem[Mohamed et~al.(2020)Mohamed, Rosca, Figurnov, and Mnih]{mohamed2020monte}
Mohamed, S., Rosca, M., Figurnov, M., and Mnih, A.
\newblock Monte carlo gradient estimation in machine learning.
\newblock \emph{Journal of Machine Learning Research}, 21\penalty0 (132):\penalty0 1--62, 2020.

\bibitem[Momeni et~al.(2023{\natexlab{a}})Momeni, Rahmani, Mall{\'e}jac, Del~Hougne, and Fleury]{momeni2023backpropagation}
Momeni, A., Rahmani, B., Mall{\'e}jac, M., Del~Hougne, P., and Fleury, R.
\newblock Backpropagation-free training of deep physical neural networks.
\newblock \emph{Science}, 382\penalty0 (6676):\penalty0 1297--1303, 2023{\natexlab{a}}.

\bibitem[Momeni et~al.(2023{\natexlab{b}})Momeni, Rahmani, Mall{\'e}jac, del Hougne, and Fleury]{momeni2023phyff}
Momeni, A., Rahmani, B., Mall{\'e}jac, M., del Hougne, P., and Fleury, R.
\newblock Phyff: Physical forward forward algorithm for in-hardware training and inference.
\newblock In \emph{Machine Learning with New Compute Paradigms}, 2023{\natexlab{b}}.

\bibitem[Nguyen \& Daugherty(2005)Nguyen and Daugherty]{nguyen2005evolutionary}
Nguyen, A.~W. and Daugherty, P.~S.
\newblock Evolutionary optimization of fluorescent proteins for intracellular fret.
\newblock \emph{Nature biotechnology}, 23\penalty0 (3):\penalty0 355--360, 2005.

\bibitem[Oguz et~al.(2024)Oguz, Dinc, Yildirim, Ke, Yoo, Wang, Yang, Moser, and Psaltis]{oguz2024optical}
Oguz, I., Dinc, N.~U., Yildirim, M., Ke, J., Yoo, I., Wang, Q., Yang, F., Moser, C., and Psaltis, D.
\newblock Optical diffusion models for image generation.
\newblock \emph{arXiv preprint arXiv:2407.10897}, 2024.

\bibitem[Ruiz et~al.(2018)Ruiz, Schulter, and Chandraker]{ruiz2018learning}
Ruiz, N., Schulter, S., and Chandraker, M.
\newblock Learning to simulate.
\newblock \emph{arXiv preprint arXiv:1810.02513}, 2018.

\bibitem[Sarkisyan et~al.(2016)Sarkisyan, Bolotin, Meer, Usmanova, Mishin, Sharonov, Ivankov, Bozhanova, Baranov, Soylemez, et~al.]{sarkisyan2016local}
Sarkisyan, K.~S., Bolotin, D.~A., Meer, M.~V., Usmanova, D.~R., Mishin, A.~S., Sharonov, G.~V., Ivankov, D.~N., Bozhanova, N.~G., Baranov, M.~S., Soylemez, O., et~al.
\newblock Local fitness landscape of the green fluorescent protein.
\newblock \emph{Nature}, 533\penalty0 (7603):\penalty0 397--401, 2016.

\bibitem[Shahriari et~al.(2015)Shahriari, Swersky, Wang, Adams, and De~Freitas]{shahriari2015taking}
Shahriari, B., Swersky, K., Wang, Z., Adams, R.~P., and De~Freitas, N.
\newblock Taking the human out of the loop: A review of bayesian optimization.
\newblock \emph{Proceedings of the IEEE}, 104\penalty0 (1):\penalty0 148--175, 2015.

\bibitem[Shirobokov et~al.(2020)Shirobokov, Belavin, Kagan, Ustyuzhanin, and Baydin]{shirobokov2020black}
Shirobokov, S., Belavin, V., Kagan, M., Ustyuzhanin, A., and Baydin, A.~G.
\newblock Black-box optimization with local generative surrogates.
\newblock \emph{Advances in neural information processing systems}, 33:\penalty0 14650--14662, 2020.

\bibitem[Si et~al.(2016)Si, Yu, and Abrahams]{si2016high}
Si, Q., Yu, R., and Abrahams, E.
\newblock High-temperature superconductivity in iron pnictides and chalcogenides.
\newblock \emph{Nature Reviews Materials}, 1\penalty0 (4):\penalty0 1--15, 2016.

\bibitem[Snoek et~al.(2012)Snoek, Larochelle, and Adams]{snoek2012practical}
Snoek, J., Larochelle, H., and Adams, R.~P.
\newblock Practical bayesian optimization of machine learning algorithms.
\newblock \emph{Advances in neural information processing systems}, 25, 2012.

\bibitem[Spall et~al.(2022)Spall, Guo, and Lvovsky]{spall2022hybrid}
Spall, J., Guo, X., and Lvovsky, A.~I.
\newblock Hybrid training of optical neural networks.
\newblock \emph{Optica}, 9\penalty0 (7):\penalty0 803--811, 2022.

\bibitem[Svanberg(1987)]{svanberg1987method}
Svanberg, K.
\newblock The method of moving asymptotes—a new method for structural optimization.
\newblock \emph{International journal for numerical methods in engineering}, 24\penalty0 (2):\penalty0 359--373, 1987.

\bibitem[Trabucco et~al.(2021)Trabucco, Kumar, Geng, and Levine]{trabucco2021conservative}
Trabucco, B., Kumar, A., Geng, X., and Levine, S.
\newblock Conservative objective models for effective offline model-based optimization.
\newblock In \emph{International Conference on Machine Learning}, pp.\  10358--10368. PMLR, 2021.

\bibitem[Wang et~al.(2018)Wang, Du, Balakrishnan, and Singh]{wang2018stochastic}
Wang, Y., Du, S., Balakrishnan, S., and Singh, A.
\newblock Stochastic zeroth-order optimization in high dimensions.
\newblock In \emph{International conference on artificial intelligence and statistics}, pp.\  1356--1365. PMLR, 2018.

\bibitem[Williams(1992)]{williams1992simple}
Williams, R.~J.
\newblock Simple statistical gradient-following algorithms for connectionist reinforcement learning.
\newblock \emph{Machine learning}, 8:\penalty0 229--256, 1992.

\bibitem[Williamson \& Trotter(2004)Williamson and Trotter]{WilliamsonTrotter2004}
Williamson, R.~E. and Trotter, H.~F.
\newblock \emph{Multivariable Mathematics}.
\newblock Pearson Prentice Hall, 4th edition, 2004.

\bibitem[Wright et~al.(2022)Wright, Onodera, Stein, Wang, Schachter, Hu, and McMahon]{wright2022deep}
Wright, L.~G., Onodera, T., Stein, M.~M., Wang, T., Schachter, D.~T., Hu, Z., and McMahon, P.~L.
\newblock Deep physical neural networks trained with backpropagation.
\newblock \emph{Nature}, 601\penalty0 (7894):\penalty0 549--555, 2022.

\bibitem[Yao et~al.(2020)Yao, Wu, Gao, Tang, Zhang, Zhang, Yang, and Qian]{yao2020fully}
Yao, P., Wu, H., Gao, B., Tang, J., Zhang, Q., Zhang, W., Yang, J.~J., and Qian, H.
\newblock Fully hardware-implemented memristor convolutional neural network.
\newblock \emph{Nature}, 577\penalty0 (7792):\penalty0 641--646, 2020.

\bibitem[Yu et~al.(2021)Yu, Ahn, Song, and Shin]{yu2021roma}
Yu, S., Ahn, S., Song, L., and Shin, J.
\newblock Roma: Robust model adaptation for offline model-based optimization.
\newblock \emph{Advances in Neural Information Processing Systems}, 34:\penalty0 4619--4631, 2021.

\bibitem[Zheng et~al.(2023)Zheng, Duan, Chen, Yang, Gao, Zhang, Xiong, and Lin]{zheng2023dual}
Zheng, Z., Duan, Z., Chen, H., Yang, R., Gao, S., Zhang, H., Xiong, H., and Lin, X.
\newblock Dual adaptive training of photonic neural networks.
\newblock \emph{Nature Machine Intelligence}, 5\penalty0 (10):\penalty0 1119--1129, 2023.

\bibitem[Zhou \& Chai(2020)Zhou and Chai]{zhou2020near}
Zhou, F. and Chai, Y.
\newblock Near-sensor and in-sensor computing.
\newblock \emph{Nature Electronics}, 3\penalty0 (11):\penalty0 664--671, 2020.

\end{thebibliography}
\bibliographystyle{icml2025}

\newpage
\appendix
\onecolumn
\section{Appendix}
\renewcommand{\theequation}{A\arabic{equation}} 
\renewcommand{\thealgorithm}{A\arabic{algorithm}} 
\setcounter{equation}{0} 
\setcounter{algorithm}{0}

\paragraph{Details of the CNON}

As mentioned in the main text (section 5.1), the CNON is described by \eqref{eq:CNO}. The coupling coefficients, \( \mathfrak{J}_{ij} \), are symmetric. 
To  enforce the condition, we construct \( \mathfrak{J}_{ij} \) from the relation:
\begin{equation}
\mathfrak{J}_{ij}  = \frac{S + S^\top}{2},
\end{equation}
where \( \top \) represents the transpose operation. Here, \( S \) is an \( N \times N \) matrix. For notational simplicity, we use a new matrix \( Q \) such that the off-diagonal elements of \( Q \) are identical to those of \( \mathfrak{J} \), and the diagonal elements are given by:
\begin{equation}
Q_{ii} = -\sum_j \mathfrak{J}_{ij}.
\end{equation}
Using this matrix, the equations of motion can be simplified into a matrix form:
\begin{equation}
\frac{d^2 \mathbf{q}_i}{dt^2} = -\sin(\pi \mathbf{q}_i) + \sum_{j=1}^N Q_{ij} \sin(\pi \mathbf{q}_j) + \mathbf{e}_i,
\end{equation}
We use the 4th order Runge–Kutta (RK4) method to solve the aforementioned ODE.

The matrix \(Q_{ij} \) is set as \( U_{ij} + Z_{ij} \), where \( U_{ij} \) and \( Z_{ij} \) are once matrix and a random matrix sampled from a uniform distribution in the range \([-1, 1]\), respectively. 
Similarly, \( \mathbf{e}_i = Z_{i} \), where \( Z_{i} \) is a random vector sampled from a uniform distribution in the range \([-1, 1]\).

\paragraph{Details of the OpAmp}

For our experiments, we simulated the two-stage operational amplifier (OpAmp) using NGspice\footnote{\url{https://ngspice.sourceforge.io/}} and the Skywater-PDK\footnote{\url{https://github.com/google/skywater-pdk}} 130nm process. The parameter ranges for the transistors were set as follows: the width and length of the biasing transistors, differential pair, load transistors, and gain transistors were varied between 2 to 32 um and 0.2 to 2 um, respectively. The width of the second stage biasing NMOS is defined as integer multiple of the width of the first stage biasing NMOS. The ratio between these two values is chosen from {1, 2, 3}. The value of the compensation cap was set between 0.1 pF and 20 pF.

\paragraph{Details of surrogate models }

We used simple MLP models as surrogate models for different tasks. Here, we provide the architecture of the MLP models in Table \ref{table:mlp_architecture}.

\begin{table*}[h]
  \centering
  \caption{Architecture of the MLP models used for the three tasks: CNON, OpAmp, and OWMS.}
  \resizebox{\textwidth}{!}{%
  \begin{tabular}{lcccccc}
  \toprule
  \textbf{Task} & \textbf{Number of Hidden Layers} & \textbf{Size of Hidden Layers} & \textbf{Activation Function} & \textbf{Bias} & \textbf{Optimizer}& \textbf{LayerNorm}  \\
  \midrule
  \textbf{CNON}  & 2                               & [256, 256]                 & GELU                          & Yes           & Adam    & No                             \\
  \textbf{OpAmp} & 4                               & [100, 100, 100, 20]          & GELU                         & Yes          & Adam           & Yes                       \\
  \textbf{OWMS}  & 5                               & [1000, 1000, 1000, 1000, 500]     & GELU                   & Yes          & Adam                 & Yes              \\
  \bottomrule
  \end{tabular}%
  }
  \label{table:mlp_architecture}
  \end{table*}

\begin{algorithm}[t!]
\footnotesize
\SetAlgoInsideSkip{2pt}
\SetAlgoSkip{2pt}
\SetInd{0.5em}{0.3em} 

\DontPrintSemicolon
\SetAlgoLined

\KwIn{%
    Dataset $\mathcal{D} = \{(\mathbf{x}_i, \mathbf{y}_i)\}_{i=1}^N$, 
    initial surrogate parameters $\boldsymbol{\theta}$, 
    learning rates $\eta_1$ and $\eta_2 > 0$, 
    number of epochs $L_{\text{epochs}}$, 
    number of nearest neighbors $K$, 
    black-box $\mathbf{F}(\cdot)$, 
    number of optimization steps $\tau$, 
    convergence threshold $\epsilon$
}
\KwOut{Trained surrogate $\hat{\mathbf{F}}(\cdot;\boldsymbol{\theta})$, optimized input $\mathbf{x}^*$}

\caption{(\textbf{Part A}) Surrogate Training with \textit{GradPIE} Loss, (\textbf{Part B}) Black-box Optimization via Trained Surrogate.}
\label{alg:gradpie_offline}

\textbf{Part A: Train the Surrogate Model.} \\
Initialize $\boldsymbol{\theta}$ ($\boldsymbol{\theta}^{(0)} \gets \boldsymbol{\theta}$) and precompute \(\mathbf{x'}_k\) for each $\mathbf{x}_i \in \mathcal{D}$. \\
\For{$l = 1$ \KwTo $L_{\text{epochs}}-1$}{
    \ForEach{mini-batch $\mathcal{B} \subseteq \mathcal{D}$}{
        Compute $\mathcal{L}_{\text{GradPIE}}$ using Eq. \eqref{eq:gradpie_loss} for $\mathcal{B}$\;
    }
    $\boldsymbol{\theta}^{(l+1)} \gets \boldsymbol{\theta}^{(l)} - \eta_1 \nabla_{\boldsymbol{\theta}}\mathcal{L}_{\text{GradPIE}}(\boldsymbol{\theta})\big|_{\boldsymbol{\theta}=\boldsymbol{\theta}^{(l)}}$ 
        \tcp*[r]{\scriptsize Update surrogate parameters}
    \If{$\mathcal{L}_{\text{GradPIE}} < \epsilon$}{
        \textbf{Break} \tcp*[r]{\scriptsize Stop if loss converges}
    }
}
Pass trained surrogate $\hat{\mathbf{F}}(\cdot; \boldsymbol{\theta})$ to Part B.

\textbf{Part B: Optimization Loop.} \\
Initialize $\mathbf{x} \sim \mathcal{N}(0, I)$ ($\mathbf{x}^{(0)} \gets \mathbf{x}$). \\
\For{$t = 1$ \KwTo $\tau-1$}{
    $\mathbf{x}^{(t+1)} \gets \mathbf{x}^{(t)} - \eta_2 \nabla_{\mathbf{x}}\hat{\mathbf{F}}(\mathbf{x}; \boldsymbol{\theta})$ 
    \tcp*[r]{\scriptsize Perform gradient descent on $\mathbf{x}$ using surrogate gradients}
}
\textbf{Return:} optimized input $\mathbf{x}^*$.
\end{algorithm}

\end{document}